\def\year{2021}\relax
\documentclass[letterpaper]{article} 
\usepackage{aaai21}  
\usepackage{times}  
\usepackage{helvet} 
\usepackage{courier}  
\usepackage[hyphens]{url}  
\usepackage{graphicx} 
\usepackage{natbib}  
\usepackage{caption} 
\usepackage[switch]{lineno}
\usepackage{multirow}
\usepackage{color}
\usepackage{amsthm,amsmath,amssymb}
\usepackage{mathrsfs}
\usepackage{algpseudocode}
\usepackage{algorithmicx,algorithm}
\usepackage{graphicx}
\usepackage{subfigure}
\usepackage{booktabs}

\definecolor{electricviolet}{rgb}{0.56, 0.0, 1.0}

\title{Meta-Curriculum Learning for Domain Adaptation \\ in Neural Machine Translation}
\author{Runzhe Zhan\thanks{Equal contribution}
        ~~Xuebo Liu$\footnotemark[1]$
        ~~Derek F. Wong\thanks{Corresponding author}
        ~~Lidia S. Chao \\}
\affiliations{
  NLP$^2$CT Lab, Department of Computer and Information Science, 
  University of Macau\\
  {nlp2ct.\{runzhe,xuebo\}@gmail.com, \{derekfw,lidiasc\}@um.edu.mo}}

\begin{document}

\maketitle

\begin{abstract}
Meta-learning has been sufficiently validated to be beneficial for low-resource neural machine translation (NMT). 
However, we find that meta-trained NMT fails to improve the translation performance of the domain unseen at the meta-training stage.
In this paper, we aim to alleviate this issue by proposing a novel {\em meta-curriculum learning} for domain adaptation in NMT.
During meta-training, the NMT first learns the similar curricula from each domain to avoid falling into a bad local optimum early, and finally learns the curricula of individualities to improve the model robustness for learning domain-specific knowledge.
Experimental results on 10 different low-resource domains show that meta-curriculum learning can improve the translation performance of both familiar and unfamiliar domains. All the codes and data are freely available at \url{https://github.com/NLP2CT/Meta-Curriculum}.
\end{abstract}

\section{Introduction}
Neural machine translation (NMT) has become the {\em de facto} method for automatic translation~\cite{sutskever2014sequence,bahdanau2014neural,vaswani2017attention}.
Despite its successes in mainstream language pairs (e.g., English to/from French) and domains (e.g., News), its potential in the translation of low-resource domains remains underexplored~\cite{koehn-knowles-2017-six}.
Meta-learning~\cite{finn2017model}, which adapts a {meta-trained} model to a new domain by using a minimal amount of training instances, becomes an intuitive appeal for improving the translation quality of low-resource domains~\cite{gu-etal-2018-meta, li2019metamt}.

\begin{figure}[t]
    \centering
    \includegraphics[scale=0.36]{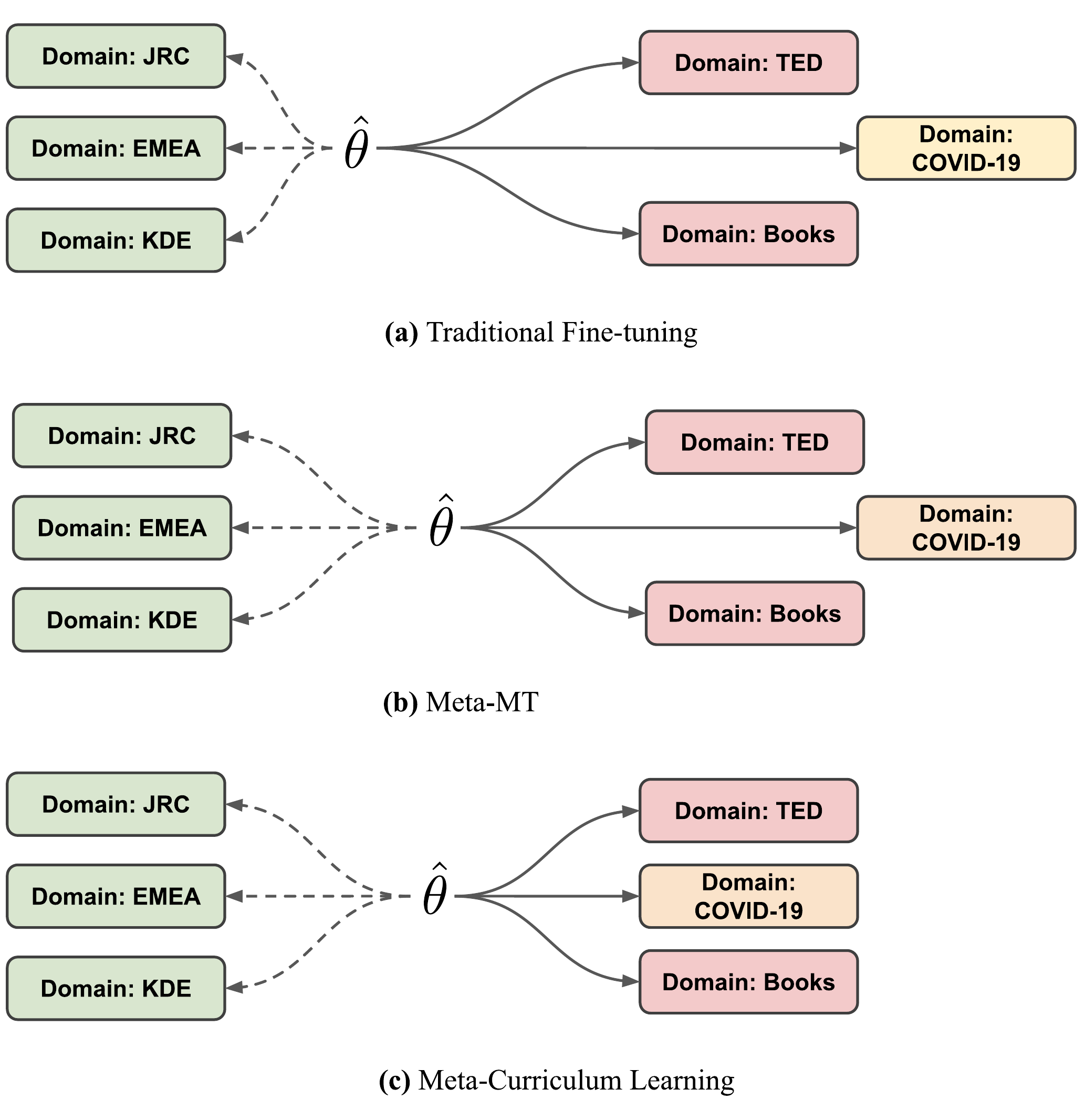}
    \caption{(a): Traditional fine-tuning favors high-resource domains (Green) over low-resource domains (Pink) and unseen domains (Orange). (b): Meta-MT partially alleviates the issue but is still biased in unseen domains. (c): Meta-curriculum learning performs well on all domains.}
    \label{fig:fig-intro}
\end{figure}

The basic idea of meta-learning for NMT is meta-training a domain-agnostic teacher model containing a set of highly robust parameters as initialization for fine-tuning domain-specific student models.
The degree of the robustness determines the performance ceilings of these student models.
\citet{sharaf2020meta} propose Meta-MT, which meta-trains the training instances of all used domains, and find that the learned initialization can generalize to these domains that improve the translation performance after individually fine-tuning.
Unfortunately, as shown in Figure~\ref{fig:fig-intro}, for those domains (e.g., COVID-19) which are not trained during meta-training, their fine-tuning performances fail to gain improvements over traditional fine-tuning.
This significantly reduces the transferability of Meta-MT, and also limits its applicability.

In this paper, we aim to alleviate the above limitations by further improving the robustness of a meta-trained teacher model.
Motivated by curriculum learning for NMT~\cite{bengio2009curriculum,cbcl,NORMCL20}, which arranges the learning curricula from easy to difficult during NMT training, we propose a novel learning strategy for low-resource domain translation, called {\em meta-curriculum learning}.
During meta-training, our proposed method learns the curricula of commonalities from different domains at the beginning, and those curricula of individualities at last.
The beginning curricula enhance the overall representation learning, while the last curricula bring forward the learning of domain-specific knowledge and thus make the meta-trained teacher more robust for fine-tuning.
Both the commonalities and individualities are measured by a language model (LM) trained on general-domain data and LMs trained on domain-specific data.

For evaluation, we first meta-train 5 domains using meta-curriculum learning, and then individually fine-tune on these 5 domains and another 5 unseen domains. 
The experimental results show that the translations of both the familiar and unfamiliar domains gain considerable improvements.
These also reveal that meta-curriculum learning can improve the transferability and applicability of meta-learning for NMT.
As far as we know, this paper is the first to integrate curriculum learning with meta-learning in NMT.

\section{Background and Related Work}
\paragraph{Neural Machine Translation (NMT)}
The sequence-to-sequence model is widely used in NMT that has the capability in directly translating source language to target language based on continuous space representation. Given a source sentence $S=(s_1,s_2,...,s_I)$, the corresponding translation produced by NMT could be denoted as $T=(t_1,t_2,...,t_J)$. The goal of the NMT model is trying to maximize the probability distribution over the target sentence $P(T|S)$, and could be formalized as the product of a series of conditional probabilities:
	\begin{equation}\label{eq_nmt}
		P(T|S)= \prod_{j=1}^J P(t_j|t_{1...j-1},S) 
	\end{equation}

Followed by the success that the attention mechanism achieved, \citet{vaswani2017attention} introduced the self-attention and multi-head attention mechanism to the prevailing encoder-decoder architecture \citep{cho2014properties} for NMT, which is named Transformer.
Compared to the previous architectures, Transformer gains a faster training speed and a better model performance.
Without loss of generality, in this paper, we choose Transformer as our testbed.

\paragraph{Domain Adaptation for NMT} The main goal of domain adaptation techniques is to narrow the domain divergence between the target tasks and training data to achieve domain-specific translation. \citet{chu2018survey} claimed that existing approaches could be categorized into two types: data-centric methods and model-centric methods. The data-centric techniques focus on maximizing the use of out-domain data to improve the translation quality \citep{currey2017copied,van2017dynamic,wang2017sentence}. The model-centric approaches can be further divided into training-oriented, architecture-oriented, and decoding-oriented methods. The most widely used strategy is continuing tuning the model using in-domain data after training the model in the general domain, which is called fine-tuning \citep{dakwale2017finetuning}, so that the distribution of the model parameters will be adapted towards the direction of the target domain.

It is also noteworthy that using curriculum learning in domain adaptation tasks can enhance model performance.
Curriculum learning is an intuitive strategy for training neural networks. It claims that the training data are supposed to be sorted by complexity or other reasonable measurement methods instead of the original random data sampling method in order to make the model be able to gradually learn from the training data within a specific order, such as sentence-level difficulty. \citet{zhang2018empirical,zhang2019curriculum} demonstrated the effectiveness of this approach in improving the domain-specific translation. The model was initialized in the general domain then fine-tuned in the in-domain data reordered by the similarity score. 
More specifically, the Moore-Lewis \citep{moore2010intelligent} and cynical data selection \citep{axelrod2017cynical} method were adopted to measure the similarity, and the probabilistic curriculum was used to form the training corpus.
In this research, a resemble Moore-Lewis approach would be adopted as the selection strategy to score the training data in curriculum learning.

\begin{figure*}[t]
    \centering
    \includegraphics[scale=0.5]{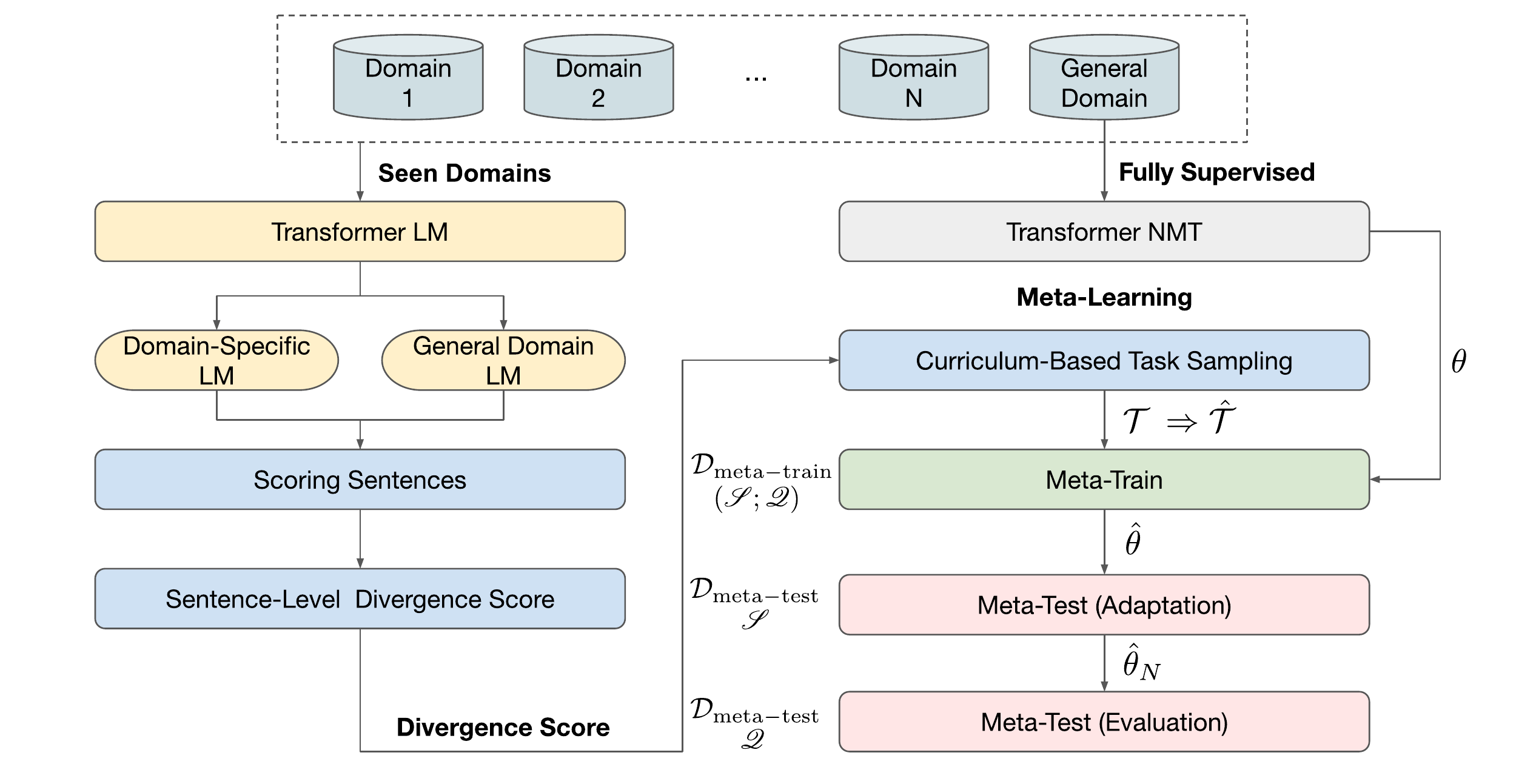}
    \caption{{\color{black}The graphical illustration of the proposed meta-curriculum {learning} algorithm. Two main parts are represented in the diagram, respectively. The process of offline scoring is illustrated on the left and the online learning part is on the right.}}
    \label{fig:fig1}
\end{figure*}

\paragraph{Meta Learning for NMT} 
{\color{black}
Since data sparsity is not only a huge challenge for machine translation but also one of the common limitations of training deep neural network models, the meta-learning algorithms~\citep{finn2017model} are proposed to obtain a robust model for better task-specific fine-tuning under low resource conditions. {Just like a human could conclude a learning pattern from previous learning experience while facing a question in the new domain}, the meta-learning aims to teach the model how to learn from diverse few-shot training exposure, which is also known as a ``learning to learn'' problem. As we all know, ordinary machine learning routinely focuses on optimizing the model $f_{\theta}$ by treating the instance $(S,T)$ of one specific domain dataset $\mathcal{D}$ as the basic unit to form the training batches. 
In contrast, the terminology ``task'' $\mathcal{T}$ whose training instances sampled from one of the training domains is the lowest learning level of meta-learning. 
Thus, the meta-training objective could be formalized as:

\begin{equation}\label{eq-metalearn}
\arg\min_\theta \mathbb{E}_{\mathcal{T}^{'}\sim p(\mathcal{T})} [\mathcal{L}_{\mathcal{T}^{'}}(f_{\theta})]
\end{equation}

Despite various meta-learning research have been conducted, the related work in the NMT field still needs to be investigated and most of them incorporate optimized-based meta-learning algorithms \citep{finn2017model,DBLP:conf/iclr/RaviL17,nichol2018first} with NMT training, especially for Model-Agnostic Meta-Learning (MAML) algorithm \citep{finn2017model}. The motivation behind optimized-based methods is training a model that could be applicable within a few optimization steps. To gain experience from simulated few-shot learning, one task $\mathcal{T}_i$ could be decomposed into two sub-sets: support set $\mathscr{S}_i$ and query set $\mathscr{Q}_i$. Typically, the support set $\mathscr{S}_i$ is used to train the parameter $\theta$, and the query set $\mathscr{Q}_i$ is adopted in evaluating the loss $\mathcal{L}({\cdot})$ using optimized parameter or testing. Therefore, the Eq.~\ref{eq-metalearn} could be further formalized in MAML algorithm:
\begin{equation}
\arg\min_\theta \sum_{\mathcal{T}_{i}} 
    {\mathcal{L}_{\mathscr{Q}_i} 
        (f_{\theta-\alpha \nabla_{\theta} \mathcal{L}(\mathscr{S}_i;\theta)})
    }
\end{equation}
where $\alpha$ is a hyper-parameter to control the pace of inner task learning. The meta-trained parameter $\hat \theta$ could be a good starting point for adapting it to various domains.

The prerequisite for MAML is that the task must be multi-domain-oriented, so multilingual translation and domain adaptation tasks are suitable to be investigated. \citet{gu-etal-2018-meta} and \citet{li2019metamt} have conducted a preliminary study on these two tasks. They improved the translation performance in the low-resource scenario by integrating the MAML algorithm with the NMT model from the perspective of lexical representation. Our work considered the most recent work, Meta-MT, which is proposed by \citet{sharaf2020meta} exploiting MAML algorithm in the domain adaptation, as a reference system in the following comparisons.
}

\section{Meta-Curriculum Learning}
The motivation of this paper is tackling the problem that the generalization ability of meta-trained model would be degraded while training within limited seen domains. 
However, re-meta-training a teacher model is costly and impractical as unseen domains are continuously increasing.
In this paper, we aim to design a novel meta-learning strategy for low-resource domain translation, which is robust enough for fine-tuning on both previously seen (meta-trained) and unseen domains.
We borrow ideas from curriculum learning~\cite{bengio2009curriculum, NORMCL20} and shared-private knowledge modeling~\cite{liu2019latent, liu-etal-2019-shared}, making the meta-training process learn curricula of different characteristics in different stages.
In the early stages, the model will sample the curricula of commonalities from all domains to avoid getting into a bad local optimum early.
While in the final stages, the curricula of individualities will be sampled for meta-training that can improve the model robustness for learning unfamiliar data.
This also tends to be beneficial for the subsequent individual fine-tuning. 
We thus call the proposed method {\em meta-curriculum learning} that combines the advantages of both techniques.

Meta-curriculum learning is formed with two parts: offline scoring and online learning. The sampling order of training instances is determined by the score given by two neural language models (LMs) which are pre-trained in the general domain and fine-tuned in the target domain, respectively. When it comes to online learning, the first-order MAML algorithm \citep{nichol2018first} is introduced as a training strategy to cooperate with reordered data in the process of online learning. The model trained using this pipeline would be adapted to other unseen domains. Figure~\ref{fig:fig1} illustrates the framework of the proposed methods. 

\subsection{Offline Scoring: LM-based Curriculum}
To make the curriculum match with the model competence, we consider making full use of model perplexity information towards the training instances. Because it is unfair to use the trained model to evaluate the instances, the LM whose architecture is similar to the Transformer decoder is suitable to be utilized in this case.

\paragraph{Sentence-level Divergence Scoring} Given an in-domain corpus $\mathcal{D}$, a general domain corpus $\mathcal{G}$, and the Transformer LM $P(\cdot)$ trained on corresponding corpus, the domain divergence $d_\mathcal{D}(S)$ of each sentence $S$ could be calculated through cross-entropy score following \citet{moore2010intelligent}:
	\begin{equation} \label{eq_sentence}
		d_\mathcal{D}(S) = H_\mathcal{D}(S)-H_\mathcal{G}(S) \\
	\end{equation}
	\begin{equation} \label{eq_nlm_sentence}
		H(S) = -\frac{\sum_{s\in S}\log P(S)}{\vert S \vert}
	\end{equation}

\noindent where $H(\cdot)$ is the token-averaged cross-entropy score generated by the Transformer LM. The higher divergence score $d(S)$ indicates that the sentence possesses more in-domain features and is more likely divergent from the samples in the general domain.

\subsection{Online Learning: Meta-Curriculum} \label{mc-learning}
\paragraph{Ordered Task Sampling}
Task sampling is an essential step for meta-learning. 
Traditional meta-learning approach uniformly samples tasks from all data, and the sentences in each task are fixed during the whole meta-training process.
Conversely, meta-curriculum {learning} samples tasks in an ordered way and the sentences in each task is changing dynamically.
At the first training steps, the tasks are mostly assembled by those sentences of commonalities in each domain, i.e., the sentences with lower divergence scores.
While at the last steps, the tasks usually consist of those sentences of individualities (i.e., higher divergence scores). 
With the increase of training steps, the average divergence score of each task also grows bigger, and the meta-learning model continuously learns difficult curricula.
To make a fair comparison, meta-curriculum {learning uses} the same data as the traditional meta-learning method used during meta-training.

\paragraph{First-order Approximation MAML}
Since we have considered the factors of data-effective meta-learning in the NMT tasks, the holistic training procedure would be proposed as following. Before performing meta-curriculum learning, the model would be firstly pre-trained on a general domain. 
Given the NMT model initialized with $\theta$ and data $\mathcal{D}$, the objective of learning could be described as $\mathrm{Learn}(f_{\theta}; \mathcal{D})$:
\begin{multline}\label{basic_NMT_learn}
	\mathop{\arg\max_{\theta}} \mathcal{L}_\mathcal{D}(f_{\theta}) = \mathop{\arg\max_{\theta}} \sum_{(S,T)\in \mathcal{D}} \log P(T|S;\theta)
\end{multline}
The traditional domain adaptation pipeline would optimize the learned parameter $\theta$ by applying a continuous training strategy in the specific in-domain corpus. However, our method applies a two-phases training policy in this process which would be divided into meta-training and meta-testing stages.
The learning algorithm of the meta-training stage as shown in Algorithm~\ref{algorithm:maml}. Followed by data sampling and reordering approach described in Section \ref{mc-learning}, it performs a traditional NMT learning process $\mathrm{Learn}(f_{\theta}; \hat{\mathscr{S}}_i)$ by adapting the model to the support set of each task $\mathcal{T}_i$, then evaluates the generalizing performance of the model parameterized with $\theta^{'}_i$ on the query set $\hat{\mathscr{Q}}_i$ of task $\mathcal{T}_i$. 
Subsequently, the loss evaluated in the downstream task would be used to update the original model parameters $\theta$. 
Furthermore, calculating second-order derivatives may cost high computation resource during training, the first-order approximation MAML (FoMAML) simplifies this step, and we use it in the implementation:
\begin{equation}\label{fomaml}
	\nabla_{\theta}\mathcal{L}_{\mathcal{T}_i}(f_{\theta_i^{'}}) \thickapprox \nabla_{\theta_i^{'}}\mathcal{L}_{\mathcal{T}_i}(f_{\theta_i^{'}})
\end{equation}

For the meta-testing stage, the optimized robust parameter $\hat \theta$ provided by the meta-training stage would be adopted to conduct a real fine-tuning operation on each domain, and it would make fast adaptation possible with few samples. 

\subsection{Overall Learning Strategy}
As stated in the previous sections, the curriculum based on the LM offline scoring and sampling strategy is combined with the meta-learning algorithm to enhance the overall representation learning. 
We reconstruct the distribution of each batch considering the unevenness of difficulty per task. 
The sentence-level divergence compared to the general domain is noteworthy to consider the curriculum of an inner task. 
Learning from the sentences with less deviation allows the model to be generalized to common knowledge, which is helpful for speeding up the convergence, while the sentences of individualities learned at the last stage make the model robust enough to learn domain-specific knowledge.

\begin{algorithm}[t]
\small
\caption{Meta-Curriculum Learning Policy} 
\hspace*{0.02in} {\bf Require:} 
	\hspace*{0.02in} Teacher model parameters $\theta$; Step size hyper-parameters $\alpha,\beta$;
\begin{algorithmic}[1]
            \For{step $m=1$ to $M$} 
            \State Prepare $I$ tasks by ordered task sampling
	\For{task $i=1$ to $I$}
		\State Evaluate $\nabla_{\theta}\mathcal{L}_{\hat{\mathcal{T}}_i}(f_{\theta})$ with support set $\hat{\mathscr{S}}_i$
		\State Conduct inner-gradient descent with parameters $\theta_i^{'} = \theta - \alpha \nabla_{\theta}\mathcal{L}_{\hat{\mathcal{T}}_i}(f_{\theta})$
	\EndFor
	\State Update $\theta \leftarrow \theta - \beta \nabla_{\theta}\sum \mathcal{L}_{\hat{\mathcal{T}}_i}(f_{\theta_i^{'}})$ with query set $\mathscr{\hat Q}$
\EndFor
\State \Return Meta-learned model parameter $\hat \theta$
\end{algorithmic}
\label{algorithm:maml}
\end{algorithm}

\section{Experiments}
The experiments are designed to investigate how the curriculum learning policy would improve the robustness of a meta-trained model and how the meta-based approach could improve the in-domain translation by contrasting with meta-learning strategy and traditional fine-tuning baselines. 
{\color{black}
More specifically, we conducted experiments on English-German (En-De) translation tasks and randomly sub-sampled $\mathcal{D}_\mathrm{meta-train}$ and $\mathcal{D}_\mathrm{meta-test}$ for each domain with specific token sizes to simulate domain adaptation tasks in a low-resource scenario.
}
The flexible NMT toolkit fairseq\footnote{https://github.com/pytorch/fairseq} and parts of code from \citet{sharaf2020meta}\footnote{https://www.dropbox.com/s/jguxb75utg1dmxl/meta-mt.zip} are used for ensuring the easily reproducing and fair comparison.

\begin{table}[t]
\small 
\centering
{\color{black}
\begin{tabular}{ccccc}
\toprule
\multirow{2}{*}{} & \multicolumn{2}{c}{$\mathcal{D}_\mathrm{meta-train}$} & \multicolumn{2}{c}{$\mathcal{D}_\mathrm{meta-test}$} \\
\cmidrule(lr){2-3} \cmidrule(lr){4-5} 
                        & \bf Support          & \bf Query         & \bf Support         & \bf Query        \\ 
\bf COVID-19                & /                & /             & 312             & 636         \\
\bf Bible                   & /                & /             & 273             & 551         \\
\bf Books                   & /                & /             & 307             & 631         \\
\bf ECB                     & /                & /             & 281             & 592         \\
\bf TED                     & /                & /             & 429             & 841         \\
\bf EMEA                    & 11,701             & 23,424         & 593             & 1,168         \\
\bf GlobalVoices            & 2,955             & 26,348         & 382             & 725         \\
\bf JRC                     & 9,734            & 19,703         & 276             & 604         \\
\bf KDE                     & 25,539             & 50,943         & 808             & 1,441         \\
\bf News                    & 11,011             & 21,895         & 352             & 693        \\
  \bottomrule 
\end{tabular}

}
\caption{\label{meta-test-info} The sampled amount of sentences used in meta-learning phases.
}
\end{table}

\subsection{Dataset}
The dataset for domain adaptation is made up of ten parallel En-De corpora (Bible-uedin \citep{christodouloupoulos2015massively}, Books, ECB, EMEA, GlobalVoices, JRC-Acquis, KDE4, TED2013, WMT-News.v2019) which are publicly available at OPUS\footnote{http://opus.nlpl.eu} \citep{tiedemann2012parallel}. The vanilla NMT was trained on the general domain with WMT-News 14 En-De shared task following the settings of~\citet{vaswani2017attention} and achieved 27.67 BLEU scores on newstest2014.
Besides, an COVID-19 En-De parallel corpus was built using authentic public institution data sources\footnote{https://www.bundesregierung.de/}$^{,}$\footnote{https://www.euro.who.int/en/health-topics/health-emergencies/coronavirus-covid-19/news/} to model a real-world domain translation.

All the sentences were filtered by the length that is no longer than $175$ (only for training and validation subset) and preprocessed with the Moses tokenizer\footnote{https://github.com/moses-smt/mosesdecoder}. 
Besides, the corpus is encoded with sentencepieces\footnote{https://github.com/google/sentencepiece} \citep{kudo-richardson-2018-sentencepiece} using $40$k joint dictionary.

The traditional meta-training dataset $\mathcal{D}_\mathrm{meta-train}$ is constructed by $160$ tasks equally sampled from five seen domains whereas the number of tasks in meta-test dataset $\mathcal{D}_\mathrm{meta-test}$ is $10$ sampled from ten domains. For each task $\mathcal{T}$, the token amount of support set $\mathscr{S}$ and query set $\mathscr{Q}$ would be approximately limited to $8$k and $16$k, respectively. Table~\ref{meta-test-info} shows the detailed statistics.

\subsection{Comparison Group}
For comparing with the existing methods of domain adaptation, the following NMT systems are included in our experiments:

\begin{itemize}
	\item \textbf{Vanilla} The system was trained on general domain (WMT-News 14 En-De) by exploiting the ordinary learning procedure without adaptation.
	\item \textbf{Traditional Fine-Tuning} Fine-tune the vanilla system following the standard continued training strategy.
	\item \textbf{Meta-MT} This system directly applies the MAML algorithm to learn from domain adaptation experience \citep{sharaf2020meta}.
	\item \textbf{Meta-Curriculum} Train the model following the online meta-learning approach guided by well-formed curriculum and then adapt it to the target domain. 
	\item \textbf{Meta-based \textit{w/o} FT} This group of experiments is to analyze the generalizability of a meta-learning system by evaluating their corresponding BLEU scores before domain adaptation and to what extent the meta-based model is affected by fine-tuning.
\end{itemize}

\subsection{Model and Learning Setup}
\paragraph{Model} The base Transformer is used as the architecture for both LM and NMT. 
Using the same network structure to evaluate the sentence can get the information about how sensitive the model is to the target sentence. 
For LM tasks, only the decoder of Transformer is used, it consists of $6$ identical decoder layers ($ d_{\rm in}=d_{\rm out}=512$, $d_{\rm hidden}=2048$, $ n_{\rm \mathrm{atten\_head}}=8$) with layer normalization. 
While resemble Transformer encoder-decoder architecture for the translation task implements the same base layer settings with $6$ identical encoders and decoders.

\paragraph{Learning}
The source side (English) LM was built for scoring the sentences, and the corresponding target sentences would be assigned by the same score and level. 
More specifically, the WMT-News Crawl 2019 English Monolingual data\footnote{http://data.statmt.org/news-crawl/en/} and full in-domain dataset (except unseen domains) are used to learn the features of general and specific domain, respectively. 
For the hyper-parameters of learning, the downstream translation task is consistent with the language modeling task. 
Both of them were trained using Adam optimizer \cite{kingma2014adam}  ($\beta_1=0.9, \beta_2=0.98$), but with different learning rates ($ lr_{\rm nlm}=5e-4, lr_{\rm finetune}=5e-5$, $lr_{\rm translation}=7e-4$, $lr_{\rm meta}=1e-5$). The learning rate scheduler and warm-up policy ($n_{\rm warmup}=4000$) for training the vanilla Transformer is the same as the \citet{vaswani2017attention} work.
Furthermore, the number of the updating epochs during the adaptation period would be strictly limited to $20$ to simulate quick adaptation and verify the robustness under limited settings.

\paragraph{Evaluation}
To fairly compare different adaptation methods, the same data source which sub-sampled as $\mathcal{D}_\mathrm{meta-test}$ would be used for fine-tuning the pre-trained model, where the support set is used for adaptation and query set is used for evaluation (the statistical information is shown in Table \ref{meta-test-info}). 
All the reported scores are case-sensitive tokenized $4$-gram BLEU~\cite{papineni-etal-2002-bleu} generated with beam size $5$. 
Besides the final fine-tuning performance of each task, the variation after adaptation is also used as the metric of the model robustness.

\begin{table*}[t]
\small
\centering
\begin{tabular}{lcccccccccc}
\toprule
& \multicolumn{5}{c}{\textbf{Unseen}}                                & \multicolumn{5}{c}{\textbf{Previously Seen (at Meta-Training Stage)}}                                                \\
\cmidrule(lr){2-6} \cmidrule(lr){7-11} 
\textbf{} &\textbf{COVID-19} & \textbf{Bible} & \textbf{Books} & \textbf{ECB}    & \textbf{TED}   & \textbf{EMEA}   & \textbf{GlobalVoices} & \textbf{JRC}   & \textbf{KDE}   & \textbf{News} \\
     \midrule
\textbf{Vanilla }  & 28.55          & 14.08          & 13.63          & 31.78          & 29.12          & 32.26          & 26.95                 & 37.15          & 24.10          & 29.50             \\
\textbf{Traditional FT} & 28.90          & 14.29          & 13.92          & 31.91          & 29.44          & 32.74          & 27.22                 & 37.40          & 25.91          & 29.44             \\
{\bf $\triangle$ FT}  & +0.35          & +0.21          & +0.29          & +0.13          & +0.32          & \textbf{+0.48} & \textbf{+0.27}        & \textbf{+0.25} & \textbf{+1.81} & -0.06             \\
     \midrule
\textbf{Meta-MT \textit{w/o} FT}  & 27.60          & 13.29          & 13.39          & 30.64          & 29.16          & 46.48          & 27.68                 & 44.70          & 31.44          & 30.91             \\
\textbf{Meta-MT} & 29.71          & 17.66          & 14.99          & 32.62          & 30.36          & 45.05          & 27.84                 & 44.01          & 31.24          & 31.17             \\
{\bf $\triangle$ MM} & +2.11          & +4.37          & +1.60           & +1.98          & \textbf{+1.20}  & -1.43          & +0.16                 & -0.69          & -0.20           & +0.26             \\
     \midrule
\textbf{Meta-Curriculum \textit{w/o} FT}	 & 27.54          & 13.52          & 13.29          & 30.98          & 29.24          & 45.50          & 28.07                 & 44.60          & 31.51          & 30.86             \\
\textbf{Meta-Curriculum}& \textbf{30.46} & \textbf{18.06} & \textbf{15.02} & \textbf{33.44} & \textbf{30.37} & \textbf{45.91} & \textbf{28.28}        & \textbf{44.85} & \textbf{31.95} & \textbf{32.00}    \\
{\bf $\triangle$ MC}& \textbf{+2.92} & \textbf{+4.54} & \textbf{+1.73} & \textbf{+2.46} & +1.13          & +0.41          & +0.21                 & \textbf{+0.25}          & +0.44          & \textbf{+1.14}        \\
\bottomrule
\end{tabular}
\caption{\label{info-table} BLEU scores over $\mathcal{D}_\mathrm{meta-test}$ query sets. \textit{w/o} denotes the system before individual adaptation. FT denotes fine-tuning. MM and MC denote Meta-MT and Meta-Curriculum, respectively. $\triangle$ markup denotes the BLEU improvement.}
\end{table*}
\section{Results and Discussion}

\paragraph{Traditional Fine-tuning} The BLEU scores shown in Table \ref{info-table} effectively confirm the superiority of the meta-based method.
In addition, to clarify the impact of domains that had been used in the meta-training stage, two sets of evaluation results are reported, \emph{Unseen} and \emph{Seen} categories. 
The results show the necessity of making this distinction. 
Due to the less prior knowledge being integrated, the traditional fine-tuning owes a huge BLEU gap on those domains seen by meta-based methods before fine-tuning, but the gap is narrowed in the unseen {groups}. 
Though far behind from the vanilla model in terms of starting BLEU points, meta-curriculum learning still outperforms the traditional method after the same {updating steps of fine-tuning} due to its excellent robustness. 
Besides, meta-based methods both show a robust and flexible tuning space of model parameters according to the performance of BLEU variation values. 
Moreover, all methods seem to encounter bottlenecks while adapting to the News domain. 
One of the reason is that the pre-trained model is also trained on this domain, thus leading to limited performance improvement with further fine-tuning.

\begin{figure}[t]
	\centering
	\includegraphics[scale=0.9]{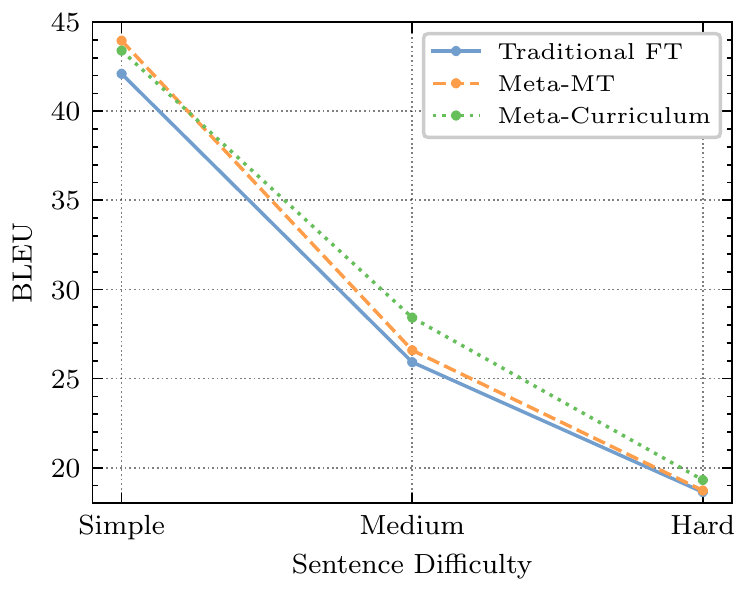}
	\caption{Performance across three types of difficulty reported on the COVID-19 domain. Meta-curriculum learning alleviates the limitation that hinders Meta-MT model from acquiring difficult knowledge.}
    \label{fig:sentence-difficulty}
\end{figure}

\paragraph{Robustness} Although the meta-trained model extracts more features from $\mathcal{D}_\mathrm{meta-train}$, it is well known that more data do not always bring a positive effect on model performance.
One of the potential pitfalls is that the distribution of model parameters tends to overfit one specific domain. 
However, the most crucial motivation of meta-learning is training a model with a robust starting point. Even if the meta-trained model did not perform well in each domain before conducting adaptation, its robustness would allow it to benefit more from fine-tuning. 
It could be interpreted as the trade-off between BLEU and robustness. 
Therefore, it is the key point we want to emphasize here because the goal of our work is improving the robustness of the meta-trained model. 

{\color{black}
As the BLEU variation ($\triangle$) shown in Table \ref{info-table}, the model trained with meta-curriculum approach is largely affected by the fine-tuning operation compared to the classical approach and Meta-MT on unseen domains. 
On the other hand, the meta-based methods show lesser improvement than the traditional fine-tuning method even overfit in some domains (EMEA, JRC, KDE) due to the fact that the meta-trained models have previously learned part of the in-domain knowledge, which somewhat limits its space for improvement. 
Despite the wide variation in the scale of the domain, the meta-trained model did not show any over-fitting signal in any domain, which rightly echoes the motivation of meta-learning. 
For example, the BLEU score it performed (27.54) was even worse than the vanilla model (28.55) before transferring to the COVID-19 domain. However, such a poor starting point has more potential to be generalized, which eventually made it outperform the baseline model in the subsequent process of adaptation. Meta-curriculum learning brings a great starting point in each domain without deviation in any field. 
}
\paragraph{Curriculum Validity}\label{sec:cval} Although the curriculum is designed from the perspective of acquisition divergence, the black-box nature of neural network makes it inconvenient to testify the scoring results. 
To verify the curriculum validity, we categorized the sentences in query set of $\mathcal{D}_\mathrm{meta-test}$ into three difficulty levels based on the scoring distribution, with each group containing nearly 200 samples. As shown in Figure~\ref{fig:sentence-difficulty}, the BLEU scores gradually decrease with increasing difficulty, which indicates that our method for designing the curriculum can distinguish adaptation difficulty.

On the other hand, from the perspective of translation quality, the meta-trained models have achieved significant improvements on the moderately difficult translation {sentences}, benefiting from its robustness. However, the Meta-MT model is likely to encounter a bottleneck in the process of acquiring more challenging knowledge, which results in worse translation performance on those {sentences} that are quite different from the general domain. Although meta-curriculum learning breaks this limitation to some extent, it is still affected by the trade-off between BLEU and robustness, making it unable to transfer some specific knowledge on extreme tasks. To investigate which kinds of bottlenecks exist in meta-learning for NMT, we perform a linguistically relevant analysis in the subsequent section.

\paragraph{Effect of Sentence Length} To investigate how meta-based methods outperform traditional fine-tuning from the perspective of linguistic, we analyze the translation results with {\em compare-mt}~\cite{neubig-etal-2019-compare}.\footnote{https://github.com/neulab/compare-mt} Meta-curriculum learning shows superior advantages on different buckets of sentence length especially for extremely long sentences, while the model trained with Meta-MT method is close to the traditional fine-tuned model. Although the Meta-MT approach is inferior to the baseline method on the most groups, it slightly outperforms the normal method in terms of translation performance on sentences {whose length is greater than $50$ tokens}, which could also prove that the meta-based method may be beneficial for the model to learn the translation of long sentences.

\begin{figure}[t]
	\centering
	\includegraphics[scale=0.9]{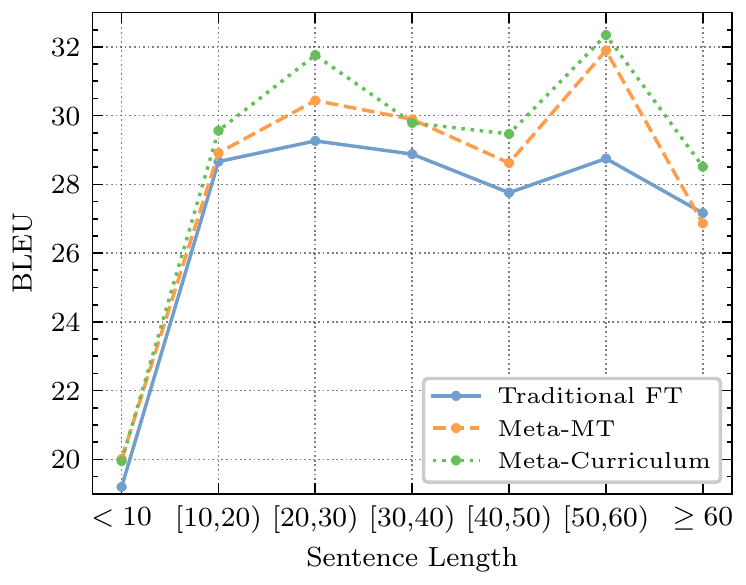}
	\caption{Performance across different buckets of sentence length reported on COVID-19 domain. The meta-based approach is clearly shows its advantages in translating long sentences.}
    \label{fig:sentence-length}
\end{figure}

\begin{figure}[t]
    \centering
    \subfigure
    {
		\includegraphics[scale=0.9]{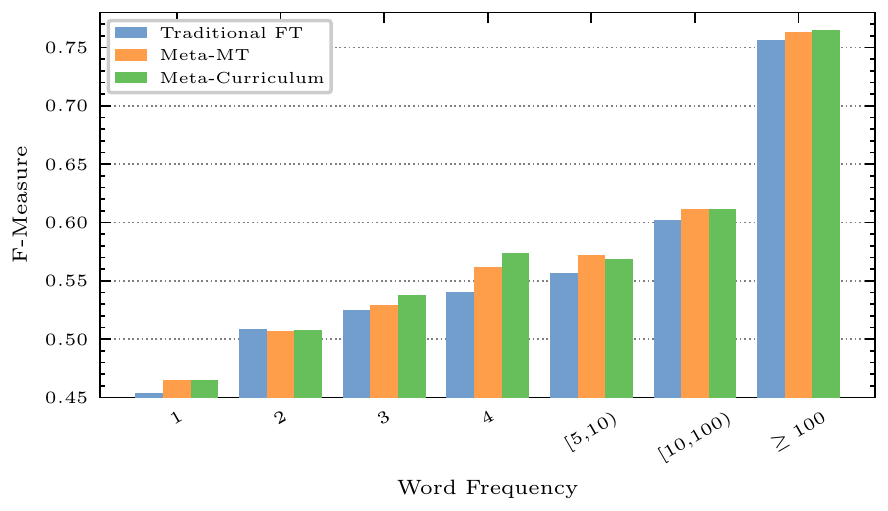}
	}
	\subfigure
	{
		\includegraphics[scale=0.9]{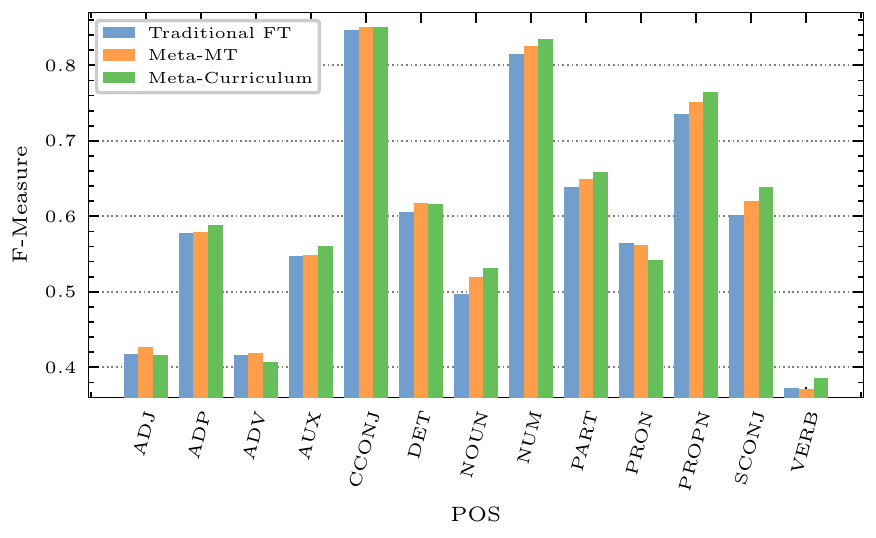}
	}
    \caption{Word-level translation across different buckets of word frequency and part-of-speech on COVID-19 domain.}
    \label{fig:word-level}
\end{figure}

\paragraph{Word-level Performance} As shown in Figure~\ref{fig:word-level}, when it comes to the word-level performance, the meta-based models do not seem to show the advantage as prominent as the sentence-level comparison, regardless of word frequency. This may be due to the fact that the meta-learning approach sacrifices the optimization space for complex lexical meanings in order to ensure the generality, which makes the lexical knowledge transfer unideal echoing the trade-off effects.
This conjecture may be corroborated by the translation accuracy of different parts-of-speech according to Figure \ref{fig:word-level}. 
Meta-curriculum approach does not make a breakthrough in some content words but performs well in terms of functional words, e.g., conjunctions. However, the correct choice for functional words may be helpful to syntactic-level translation which may explain why the meta-based models are skillful at translating long sentences. 

Furthermore, there is still one possible reason for this phenomenon. 
As pointed by \citet{gu-etal-2018-meta}, the effectiveness of the meta-learning algorithm in NMT tasks would be affected by both the fertility of meta-training knowledge and its relevance to the target domain. 
{For example,} the COVID-19 domain, not only its knowledge distance from meta-training is relatively far, but also the in-domain knowledge provided in the meta-testing is poor, leading to limited improvement in this case.
Therefore, in order to improve the robustness of the meta-based approach, it is advisable to enhance and correct the lexical meanings of the corresponding domains under low-resource {scenario} for making the translation quality produced by the meta-trained system better.

\section{Conclusion and Future Works}
We propose a novel meta-curriculum learning for domain adaptation in NMT, which improves the robustness of the meta-trained domain-agnostic teacher model and the fine-tuning performances of the domain-specific student models.
Unlike existing meta-learning methods for NMT which only improve the translations of the domains previously seen during the meta-training, meta-curriculum learning can also enhance the representation learning of those unseen domains, especially those instances of individualities.
This results in an overall performance boost in terms of BLEU scores, and consistently better translations over sentences of different lengths, words of different frequencies and some part-of-speech tags.
In the future, we would like to add curriculum learning methods into the fine-tuning process to further enhance meta-learning for NMT.

\section{Acknowledgments}
This work was supported in part by the National Natural Science Foundation of China (Grant No. 61672555), the Joint Project of the Science and Technology Development Fund, Macau SAR and National Natural Science Foundation of China (Grant No. FDCT/0072/2021/AFJ), the Science and Technology Development Fund, Macau SAR (Grant No. 0101/2019/A2), and the Multi-year Research Grant from the University of Macau (Grant No. MYRG2020-00054-FST). We thank the anonymous reviewers for their insightful comments.

\bibliography{aaai21}

\begin{thebibliography}{31}
\providecommand{\natexlab}[1]{#1}
\providecommand{\url}[1]{\texttt{#1}}
\providecommand{\urlprefix}{URL }
\expandafter\ifx\csname urlstyle\endcsname\relax
  \providecommand{\doi}[1]{doi:\discretionary{}{}{}#1}\else
  \providecommand{\doi}{doi:\discretionary{}{}{}\begingroup
  \urlstyle{rm}\Url}\fi

\bibitem[{Axelrod(2017)}]{axelrod2017cynical}
Axelrod, A. 2017.
\newblock Cynical selection of language model training data.
\newblock \emph{arXiv preprint arXiv:1709.02279} .

\bibitem[{Bahdanau, Cho, and Bengio(2015)}]{bahdanau2014neural}
Bahdanau, D.; Cho, K.; and Bengio, Y. 2015.
\newblock Neural Machine Translation by Jointly Learning to Align and
  Translate.
\newblock In \emph{3rd International Conference on Learning Representations,
  {ICLR} 2015, Conference Track Proceedings}.

\bibitem[{Bengio et~al.(2009)Bengio, Louradour, Collobert, and
  Weston}]{bengio2009curriculum}
Bengio, Y.; Louradour, J.; Collobert, R.; and Weston, J. 2009.
\newblock Curriculum learning.
\newblock In \emph{Proceedings of the 26th annual international conference on
  machine learning}, 41--48.

\bibitem[{Cho et~al.(2014)Cho, van Merri{\"e}nboer, Bahdanau, and
  Bengio}]{cho2014properties}
Cho, K.; van Merri{\"e}nboer, B.; Bahdanau, D.; and Bengio, Y. 2014.
\newblock On the Properties of Neural Machine Translation: Encoder--Decoder
  Approaches.
\newblock In \emph{Proceedings of SSST-8, Eighth Workshop on Syntax, Semantics
  and Structure in Statistical Translation}, 103--111.

\bibitem[{Christodouloupoulos and
  Steedman(2015)}]{christodouloupoulos2015massively}
Christodouloupoulos, C.; and Steedman, M. 2015.
\newblock A massively parallel corpus: the Bible in 100 languages.
\newblock \emph{Language resources and evaluation} 49(2): 375--395.

\bibitem[{Chu and Wang(2018)}]{chu2018survey}
Chu, C.; and Wang, R. 2018.
\newblock A Survey of Domain Adaptation for Neural Machine Translation.
\newblock In \emph{Proceedings of the 27th International Conference on
  Computational Linguistics}, 1304--1319.

\bibitem[{Currey, Miceli-Barone, and Heafield(2017)}]{currey2017copied}
Currey, A.; Miceli-Barone, A.~V.; and Heafield, K. 2017.
\newblock Copied monolingual data improves low-resource neural machine
  translation.
\newblock In \emph{Proceedings of the Second Conference on Machine
  Translation}, 148--156.

\bibitem[{Dakwale and Monz(2017)}]{dakwale2017finetuning}
Dakwale, P.; and Monz, C. 2017.
\newblock Finetuning for neural machine translation with limited degradation
  across in-and out-of-domain data.
\newblock \emph{Proceedings of the XVI Machine Translation Summit} 117.

\bibitem[{Finn, Abbeel, and Levine(2017)}]{finn2017model}
Finn, C.; Abbeel, P.; and Levine, S. 2017.
\newblock Model-agnostic meta-learning for fast adaptation of deep networks.
\newblock In \emph{Proceedings of the 34th International Conference on Machine
  Learning-Volume 70}, 1126--1135.

\bibitem[{Gu et~al.(2018)Gu, Wang, Chen, Li, and Cho}]{gu-etal-2018-meta}
Gu, J.; Wang, Y.; Chen, Y.; Li, V. O.~K.; and Cho, K. 2018.
\newblock Meta-Learning for Low-Resource Neural Machine Translation.
\newblock In \emph{Proceedings of the 2018 Conference on Empirical Methods in
  Natural Language Processing}, 3622--3631. Association for Computational
  Linguistics.

\bibitem[{Kingma and Ba(2015)}]{kingma2014adam}
Kingma, D.~P.; and Ba, J. 2015.
\newblock Adam: {A} Method for Stochastic Optimization.
\newblock In \emph{3rd International Conference on Learning Representations,
  Conference Track Proceedings}.

\bibitem[{Koehn and Knowles(2017)}]{koehn-knowles-2017-six}
Koehn, P.; and Knowles, R. 2017.
\newblock Six Challenges for Neural Machine Translation.
\newblock In \emph{Proceedings of the First Workshop on Neural Machine
  Translation}, 28--39. Association for Computational Linguistics.

\bibitem[{Kudo and Richardson(2018)}]{kudo-richardson-2018-sentencepiece}
Kudo, T.; and Richardson, J. 2018.
\newblock {S}entence{P}iece: A simple and language independent subword
  tokenizer and detokenizer for Neural Text Processing.
\newblock In \emph{Proceedings of the 2018 Conference on Empirical Methods in
  Natural Language Processing: System Demonstrations}, 66--71.

\bibitem[{Li, Wang, and Yu(2020)}]{li2019metamt}
Li, R.; Wang, X.; and Yu, H. 2020.
\newblock MetaMT, a Meta Learning Method Leveraging Multiple Domain Data for
  Low Resource Machine Translation.
\newblock In \emph{The Thirty-Fourth {AAAI} Conference on Artificial
  Intelligence, 2020}, 8245--8252. {AAAI} Press.

\bibitem[{Liu et~al.(2020)Liu, Lai, Wong, and Chao}]{NORMCL20}
Liu, X.; Lai, H.; Wong, D.~F.; and Chao, L.~S. 2020.
\newblock Norm-Based Curriculum Learning for Neural Machine Translation.
\newblock In \emph{Proceedings of the 58th Annual Meeting of the Association
  for Computational Linguistics}, 427--436. Association for Computational
  Linguistics.

\bibitem[{Liu et~al.(2019{\natexlab{a}})Liu, Wong, Chao, and
  Liu}]{liu2019latent}
Liu, X.; Wong, D.~F.; Chao, L.~S.; and Liu, Y. 2019{\natexlab{a}}.
\newblock Latent Attribute Based Hierarchical Decoder for Neural Machine
  Translation.
\newblock \emph{IEEE/ACM Transactions on Audio, Speech, and Language
  Processing} 27(12): 2103--2112.

\bibitem[{Liu et~al.(2019{\natexlab{b}})Liu, Wong, Liu, Chao, Xiao, and
  Zhu}]{liu-etal-2019-shared}
Liu, X.; Wong, D.~F.; Liu, Y.; Chao, L.~S.; Xiao, T.; and Zhu, J.
  2019{\natexlab{b}}.
\newblock Shared-Private Bilingual Word Embeddings for Neural Machine
  Translation.
\newblock In \emph{Proceedings of the 57th Annual Meeting of the Association
  for Computational Linguistics}, 3613--3622. Association for Computational
  Linguistics.

\bibitem[{Moore and Lewis(2010)}]{moore2010intelligent}
Moore, R.~C.; and Lewis, W. 2010.
\newblock Intelligent selection of language model training data.
\newblock In \emph{Proceedings of the {ACL} 2010 Conference Short Papers},
  220--224. Association for Computational Linguistics.

\bibitem[{Neubig et~al.(2019)Neubig, Dou, Hu, Michel, Pruthi, and
  Wang}]{neubig-etal-2019-compare}
Neubig, G.; Dou, Z.-Y.; Hu, J.; Michel, P.; Pruthi, D.; and Wang, X. 2019.
\newblock compare-mt: A Tool for Holistic Comparison of Language Generation
  Systems.
\newblock In \emph{Proceedings of the 2019 Conference of the North {A}merican
  Chapter of the Association for Computational Linguistics (Demonstrations)},
  35--41. Minneapolis, Minnesota: Association for Computational Linguistics.

\bibitem[{Nichol, Achiam, and Schulman(2018)}]{nichol2018first}
Nichol, A.; Achiam, J.; and Schulman, J. 2018.
\newblock On first-order meta-learning algorithms.
\newblock \emph{arXiv preprint arXiv:1803.02999} .

\bibitem[{Papineni et~al.(2002)Papineni, Roukos, Ward, and
  Zhu}]{papineni-etal-2002-bleu}
Papineni, K.; Roukos, S.; Ward, T.; and Zhu, W.-J. 2002.
\newblock BLEU: a method for automatic evaluation of machine translation.
\newblock In \emph{Proceedings of the 40th annual meeting of the Association
  for Computational Linguistics}, 311--318.

\bibitem[{Platanios et~al.(2019)Platanios, Stretcu, Neubig, P{\'{o}}czos, and
  Mitchell}]{cbcl}
Platanios, E.~A.; Stretcu, O.; Neubig, G.; P{\'{o}}czos, B.; and Mitchell,
  T.~M. 2019.
\newblock Competence-based Curriculum Learning for Neural Machine Translation.
\newblock In \emph{Proceedings of the 2019 Conference of the North {A}merican
  Chapter of the Association for Computational Linguistics: Human Language
  Technologies, Volume 1 (Long and Short Papers)}, 1162--1172. Association for
  Computational Linguistics.

\bibitem[{Ravi and Larochelle(2017)}]{DBLP:conf/iclr/RaviL17}
Ravi, S.; and Larochelle, H. 2017.
\newblock Optimization as a Model for Few-Shot Learning.
\newblock In \emph{5th International Conference on Learning Representations,
  {ICLR} 2017, Toulon, France, April 24-26, 2017, Conference Track
  Proceedings}.

\bibitem[{Sharaf, Hassan, and Daum{\'e}~III(2020)}]{sharaf2020meta}
Sharaf, A.; Hassan, H.; and Daum{\'e}~III, H. 2020.
\newblock Meta-Learning for Few-Shot {NMT} Adaptation.
\newblock In \emph{Proceedings of the Fourth Workshop on Neural Generation and
  Translation}, 43--53. Association for Computational Linguistics.

\bibitem[{Sutskever, Vinyals, and Le(2014)}]{sutskever2014sequence}
Sutskever, I.; Vinyals, O.; and Le, Q.~V. 2014.
\newblock Sequence to Sequence Learning with Neural Networks.
\newblock In \emph{Advances in Neural Information Processing Systems 27},
  3104--3112. Curran Associates, Inc.

\bibitem[{Tiedemann(2012)}]{tiedemann2012parallel}
Tiedemann, J. 2012.
\newblock Parallel Data, Tools and Interfaces in OPUS.
\newblock In \emph{Proceedings of the Eight International Conference on
  Language Resources and Evaluation (LREC'12)}, volume 2012, 2214--2218.
  European Language Resources Association (ELRA).

\bibitem[{van~der Wees, Bisazza, and Monz(2017)}]{van2017dynamic}
van~der Wees, M.; Bisazza, A.; and Monz, C. 2017.
\newblock Dynamic Data Selection for Neural Machine Translation.
\newblock In \emph{Proceedings of the 2017 Conference on Empirical Methods in
  Natural Language Processing}, 1400--1410.

\bibitem[{Vaswani et~al.(2017)Vaswani, Shazeer, Parmar, Uszkoreit, Jones,
  Gomez, Kaiser, and Polosukhin}]{vaswani2017attention}
Vaswani, A.; Shazeer, N.; Parmar, N.; Uszkoreit, J.; Jones, L.; Gomez, A.~N.;
  Kaiser, {\L}.; and Polosukhin, I. 2017.
\newblock Attention is all you need.
\newblock In \emph{Advances in neural information processing systems},
  5998--6008.

\bibitem[{Wang et~al.(2017)Wang, Finch, Utiyama, and Sumita}]{wang2017sentence}
Wang, R.; Finch, A.; Utiyama, M.; and Sumita, E. 2017.
\newblock Sentence embedding for neural machine translation domain adaptation.
\newblock In \emph{Proceedings of the 55th Annual Meeting of the Association
  for Computational Linguistics (Volume 2: Short Papers)}, 560--566.

\bibitem[{Zhang et~al.(2018)Zhang, Kumar, Khayrallah, Murray, Gwinnup,
  Martindale, McNamee, Duh, and Carpuat}]{zhang2018empirical}
Zhang, X.; Kumar, G.; Khayrallah, H.; Murray, K.; Gwinnup, J.; Martindale,
  M.~J.; McNamee, P.; Duh, K.; and Carpuat, M. 2018.
\newblock An empirical exploration of curriculum learning for neural machine
  translation.
\newblock \emph{arXiv preprint arXiv:1811.00739} .

\bibitem[{Zhang et~al.(2019)Zhang, Shapiro, Kumar, McNamee, Carpuat, and
  Duh}]{zhang2019curriculum}
Zhang, X.; Shapiro, P.; Kumar, G.; McNamee, P.; Carpuat, M.; and Duh, K. 2019.
\newblock Curriculum Learning for Domain Adaptation in Neural Machine
  Translation.
\newblock In \emph{Proceedings of the 2019 Conference of the North American
  Chapter of the Association for Computational Linguistics: Human Language
  Technologies, Volume 1 (Long and Short Papers)}, 1903--1915.

\end{thebibliography}

\end{document}